\documentclass[11pt]{article}
\usepackage{semeval2014}
\usepackage{times}
\usepackage{url}
\usepackage{latexsym}
\usepackage{amssymb}
\usepackage{afterpage}

\title{SemEval-2014 Task 9: Sentiment Analysis in Twitter}

\author{
{\bf Sara Rosenthal }\\Columbia University\\{\small {\tt sara@cs.columbia.edu}}\\
{\bf Preslav Nakov}\\Qatar Computing Research Institute\\{\small {\tt pnakov@qf.org.qa}}\\
\And
{\bf Alan Ritter}\\Carnegie Mellon University\\{\small {\tt rittera@cs.cmu.edu}}\\
{\bf Veselin Stoyanov}\\Johns Hopkins University\\{\small {\tt ves@cs.jhu.edu}}\\
}

\date{}

\begin{document}
\maketitle
\begin{abstract}
We describe the \emph{Sentiment Analysis in Twitter} task, ran as part of SemEval-2014. It is a continuation of the last year's task that ran successfully as part of SemEval-2013. As in 2013, this was the most popular SemEval task; a total of 46 teams contributed 27 submissions for subtask A (21 teams) and 50 submissions for subtask B (44 teams). This year, we introduced three new test sets: (\emph{i}) regular tweets, (\emph{ii}) sarcastic tweets, and (\emph{iii}) LiveJournal sentences. We further tested on (\emph{iv}) 2013 tweets, and (\emph{v}) 2013 SMS messages.
The highest F1-score on (\emph{i}) was achieved by \emph{NRC-Canada} at 86.63 for subtask A and by TeamX at 70.96 for subtask B.
\end{abstract}

\section{Introduction}

In the past decade, new forms of communication have emerged and have become ubiquitous through social media.  Microblogs (e.g., Twitter), Weblogs (e.g., LiveJournal) and cell phone messages (SMS) are often used to share opinions and sentiments about the surrounding world, and the availability of social content generated on sites such as Twitter creates new opportunities to automatically study public opinion.\blfootnote{This work is licensed under a Creative Commons Attribution 4.0 International Licence. Page numbers and proceedings footer are added by the organisers. Licence details: http://creativecommons.org/licenses/by/4.0/}

Working with these informal text genres presents new challenges for natural language processing
beyond those encountered when working with more traditional text genres such as newswire.
The language in social media is very informal, with creative spelling and punctuation, misspellings, slang, new words, URLs, and genre-specific terminology and abbreviations, e.g., RT for re-tweet and \#hashtags\footnote{Hashtags are a type of tagging for Twitter messages.}.

Moreover, tweets and SMS messages are short: a sentence or a headline rather than a document.

How to handle such challenges so as to automatically mine and understand people's opinions and sentiments
has only recently been the subject of research \cite{Jansen09,Barbosa10,Bifet11,Davidov10,oconnor10,Pak10,Tumasjan10,Kouloumpis11}.

Several corpora with detailed opinion and sentiment annotation have been made freely available,
e.g., the MPQA newswire corpus \cite{Wiebe05},
the movie reviews corpus \cite{Pang:2002:TUS},
or the restaurant and laptop reviews corpora that are part of this year's SemEval Task 4 \cite{Semeval2014task4}.
These corpora have proved very valuable as resources for learning about the language of sentiment in general, but they do not focus on tweets. While some Twitter sentiment datasets were created prior to SemEval-2013, they were either small and proprietary, such as the i-sieve corpus \cite{Kouloumpis11} or focused solely on message-level sentiment.

Thus, the primary goal of our SemEval task is to promote research
that will lead to better understanding of how sentiment is conveyed in Social Media.
Toward that goal, we created the SemEval Tweet corpus as part of our inaugural Sentiment Analysis in Twitter Task, SemEval-2013 Task 2 \cite{semeval2013}. It contains tweets and SMS messages with sentiment expressions annotated with contextual phrase-level and message-level polarity. This year, we extended the corpus by adding new tweets and LiveJournal sentences.

Another interesting phenomenon that has been studied in Twitter is the use of the \#sarcasm hashtag to indicate that a tweet should not be taken literally~\cite{conf/acl/Gonzalez-IbanezMW11,liebrecht-kunneman-vandenbosch:2013:WASSA}.
In fact, sarcasm indicates that the message polarity should be flipped.
With this in mind, this year, we also evaluate on sarcastic tweets.

In the remainder of this paper, we first describe the task, the dataset creation process and the evaluation methodology.
We then summarize the characteristics of the approaches taken by the participating systems, and we discuss their scores.

\section{Task Description}

As SemEval-2013 Task 2, we included two subtasks: an expression-level subtask and a message-level subtask.
Participants could choose to participate in either or both.
Below we provide short descriptions of the objectives of these two subtasks.

\begin{description}
  \item[Subtask A: Contextual Polarity Disambiguation] Given a message containing a marked instance of a word or a phrase, determine whether that instance is positive, negative or neutral in that context.
      The instance boundaries were provided:
      this was a classification task, not an entity recognition task.
  \item [Subtask B: Message Polarity Classification]
  Given a message, decide whether it is of positive, negative, or neutral sentiment.
    For messages conveying both positive and negative sentiment, the stronger one is to be chosen.
\end{description}

Each participating team was allowed to submit results for two different systems per subtask: one constrained, and one unconstrained.
A constrained system could only use the provided data for training, but it could also use other resources such as lexicons obtained elsewhere.
An unconstrained system could use any additional data as part of the training process;
this could be done in a supervised, semi-supervised, or unsupervised fashion.

Note that constrained/unconstrained refers to the data used to train a classifier. For example, if other data (excluding the test data) was used to develop a sentiment lexicon, and the lexicon was used to generate features, the system would still be constrained. However, if other data (excluding the test data) was used to develop a sentiment lexicon, and this lexicon was used to automatically label additional Tweet/SMS messages and then used with the original data to train the classifier, then such a system would be considered unconstrained.

\section{Datasets}

 \begin{table}[ht]
\small
\begin{center}
\begin{tabular}{|l |r |r | r |}
\hline
\multicolumn{1}{|c|}{\bf Corpus} & \bf Positive & \bf Negative & \bf Objective \\
&  &  & \bf / Neutral \\
\hline
Twitter2013-train  & 5,895 & 3,131 &  471  \\
Twitter2013-dev  & 648 & 430 &  57    \\
Twitter2013-test  & 2,734 & 1,541 & 160  \\
SMS2013-test  & 1,071 & 1,104 & 159 \\
\hline
Twitter2014-test  & 1,807 & 578 & 88 \\
Twitter2014-sarcasm & 82 & 37 & 5 \\
LiveJournal2014-test  & 660 & 511 & 144 \\
\hline
\end{tabular}
\caption{Dataset statistics for Subtask A.}
\label{T:CorpusStatsA}
\end{center}
\end{table}

In this section, we describe the process of collecting and annotating
the 2014 testing tweets, including the sarcastic ones, and LiveJournal sentences.

 \subsection{Datasets Used}

For training and development, we released the Twitter train/dev/test datasets from SemEval-2013 task 2,
as well as the SMS test set, which uses messages from the NUS SMS corpus \cite{SMScorpus},
which we annotated for sentiment in 2013.

We further added a new 2014 Twitter test set, as well as a small set of tweets that contained the \#sarcasm hashtag to determine how sarcasm affects the tweet polarity. Finally, we included sentences from LiveJournal in order to determine how systems trained on Twitter perform on other sources. The statistics for each dataset and for each subtask are shown in Tables~\ref{T:CorpusStatsA} and ~\ref{T:CorpusStatsB}.

\begin{table}[ht]
\small
\begin{center}
\begin{tabular}{|l |r |r | r |}
\hline
\multicolumn{1}{|c|}{\bf Corpus} & \bf Positive & \bf Negative & \bf Objective \\
&  &  & \bf / Neutral \\
\hline
Twitter2013-train & 3,662 & 1,466 &  4,600  \\
Twitter2013-dev & 575 & 340 &  739  \\
Twitter2013-test  & 1,572 & 601 & 1,640  \\
SMS2013-test  & 492 & 394 & 1,207 \\
\hline
Twitter2014-test  & 982 & 202 & 669 \\
Twitter2014-sarcasm & 33 & 40 & 13 \\
LiveJournal2014-test  & 427 & 304 & 411 \\
\hline
\end{tabular}
\caption{Dataset statistics for Subtask B.}
\label{T:CorpusStatsB}
\end{center}
\end{table}

 \subsection{Annotation}

\begin{table*}[t]
\small
\begin{center}
\begin{tabular}{|l |p{11cm}|l|}
\hline
\multicolumn{1}{|c|}{\bf Source} & \multicolumn{1}{|c|}{\bf Example} & \bf Polarity \\
\hline
Twitter & Why would you [still]- wear shorts when it's this cold?!  I [love]+ how Britain see's a bit of sun and they're [like 'OOOH]+ LET'S STRIP!' & positive \\
\hline
SMS & [Sorry]- I think tonight [cannot]- and I [not feeling well]- after my rest. & negative \\
\hline
LiveJournal & [Cool]+ posts , dude ; very [colorful]+ , and [artsy]+ . & positive \\
\hline
Twitter Sarcasm & [Thanks]+ manager for putting me on the schedule for Sunday & negative \\
\hline
\end{tabular}
\caption{Example of polarity for each source of messages.
        The target phrases are marked in [$\ldots$], and are followed by their polarity;
        the sentence-level polarity is shown in the last column.}
\label{T:Examples}
\end{center}
\end{table*}

We annotated the new tweets as in 2013:
by identifying tweets from popular topics that contain sentiment-bearing words by using SentiWordNet \cite{swn} as a filter. We altered the annotation task for the sarcastic tweets,
displaying them to the Mechanical Turk annotators without the \#sarcasm hashtag; the Turkers had to determine whether the tweet is sarcastic on their own. Moreover, we asked Turkers to indicate the degree of sarcasm as (a)~definitely sarcastic, (b)~probably sarcastic, and (c)~not sarcastic.

As in 2013, we combined the annotations using intersection,
where a word had to appear in 2/3 of the annotations to be accepted.
An annotated example from each source is shown in Table~\ref{T:Examples}.

\subsection{Tweets Delivery}

We did not deliver the annotated tweets to the participants directly;
instead, we released annotation indexes, a list of corresponding Twitter IDs,
and a download script that extracts the corresponding tweets via the Twitter API.\footnote{https://dev.twitter.com}
We provided the tweets in this manner in order to ensure that Twitter's terms of service are not violated. Unfortunately, due to this restriction, the task participants had access to different number of training tweets
depending on when they did the downloading.
This varied between a minimum of 5,215 tweets and the full set of 10,882 tweets.
On average the teams were able to collect close to 9,000 tweets;
for teams that did not participate in 2013, this was about 8,500.
The difference in training data size did not seem to have had a major impact.
In fact, the top two teams in subtask B (coooolll and TeamX) trained on less than 8,500 tweets.

\section{Scoring}
\label{Section:scoring}

The participating systems were required to perform a three-way classification for both subtasks.
A particular marked phrase (for subtask A) or an entire message (for subtask B)
was to be classified as {\em positive}, {\em negative} or {\em objective/neutral}.
We scored the systems by computing a score for predicting positive/negative phrases/messages.
For instance, to compute positive precision, $p_{pos}$, we find the number of phrases/messages that a system correctly predicted to be positive,
and we divide that number by the total number it predicted to be positive.
To compute positive recall, $r_{pos}$, we find the number of phrases/messages correctly predicted to be positive and we divide that number by the total number of positives in the gold standard.
We then calculate F1-score for the positive class
as follows $F_{pos}=\frac{2(p_{pos}+r_{pos})}{p_{pos}*r_{pos}}$. We carry out a similar computation for $F_{neg}$,
for the negative phrases/messages.
The overall score
is then
$F=(F_{pos}+F_{neg})/2$.

We used the two test sets from 2013 and the three from 2014,
which we combined into one test set
and we shuffled to make it hard to guess which set a sentence came from.
This guaranteed that participants would submit predictions for all five test sets.
It also allowed us to test how well systems trained on standard tweets generalize to sarcastic tweets and to LiveJournal sentences, without the participants putting extra efforts into this.
The participants were also not informed about the source the extra test sets come from.

We provided the participants with a scorer that outputs the overall score $F$
and a confusion matrix for each of the five test sets.

\section{Participants and Results}


The results are shown in Tables \ref{tab:SubtaskA} and \ref{tab:SubtaskB},
and the team affiliations are shown in Table \ref{T:teams}.
Tables \ref{tab:SubtaskA} and \ref{tab:SubtaskB} contain results on the two progress test sets (tweets and SMS messages), which are the official test sets from the 2013 edition of the task, and on the three new official 2014 testsets (tweets, tweets with sarcasm, and LiveJournal). The tables further show macro- and micro-averaged results over the 2014 datasets. There is an index for each result showing the relative rank of that result within the respective column.
The participating systems are ranked by their score on the Twitter-2014 testset, which is the official ranking for the task;
all remaining rankings are secondary.


As we mentioned above, the participants were not told that the 2013 test sets would be included in the big 2014 test set,
so that they do not overtune their systems on them.
However, the 2013 test sets were made available for development,
but it was explicitly forbidden to use them for training.
Still, some participants did not notice this restriction, which resulted in their unusually high scores on Twitter2013-test;
we did our best to identify all such cases, and we asked the authors to submit corrected runs.
The tables mark such resubmissions accordingly.

Most of the submissions were constrained, with just a few unconstrained:
7 out of 27 for subtask A, and 8 out of 50 for subtask B.
In any case, the best systems were constrained.
Some teams participated with both a constrained and an unconstrained system,
but the unconstrained system was not always better than the constrained one:
sometimes it was worse, sometimes it performed the same.
Thus, we decided to produce a single ranking, including both constrained and unconstrained systems,
where we mark the latter accordingly.

\subsection{Subtask A}

\begin{table*}[t!]
    \centering
    \begin{small}
    \begin{tabular}{|c|l|c|cc|ccc|cc|}
    \hline
   & & \bf Uncon-& \multicolumn{2}{c|}{\bf 2013: Progress} & \multicolumn{3}{c|}{\bf 2014: Official} & \multicolumn{2}{c|}{\bf 2014: Average}\\
   \bf \# & \multicolumn{1}{|c|}{\bf System} & \bf strain.? & \bf Tweet & \bf SMS & \bf Tweet & \bf Tweet & \bf Live- & \bf Macro & \bf Micro\\
   & &  & &  & \bf & \bf sarcasm & \bf Journal & & \\
    \hline
    1 &  NRC-Canada  &   & 90.14$_{1}$ & 88.03$_{4}$ & 86.63$_{1}$ & 77.13$_{5}$ & 85.49$_{2}$ & 83.08$_{2}$ & 85.61$_{1}$ \\
    2 &  SentiKLUE  &   & 90.11$_{2}$ & 85.16$_{8}$ & 84.83$_{2}$ & 79.32$_{3}$ & 85.61$_{1}$ & 83.25$_{1}$ & 85.15$_{2}$ \\
    3 &  CMUQ-Hybrid$^{*}$  &   & 88.94$_{4}$ & 87.98$_{5}$ & 84.40$_{3}$ & 76.99$_{6}$ & 84.21$_{3}$ & 81.87$_{3}$ & 84.05$_{3}$ \\
    4 &  CMU-Qatar$^{*}$  &   & 89.85$_{3}$ & 88.08$_{3}$ & 83.45$_{4}$ & 78.07$_{4}$ & 83.89$_{5}$ & 81.80$_{4}$ & 83.56$_{4}$ \\
    5 &  ECNU &  \checkmark  & 87.29$_{6}$ & 89.26$_{2}$ & 82.93$_{5}$ & 73.71$_{8}$ & 81.69$_{7}$ & 79.44$_{7}$ & 81.85$_{6}$ \\
    6 &  ECNU  &   & 87.28$_{7}$ & 89.31$_{1}$ & 82.67$_{6}$ & 73.71$_{9}$ & 81.67$_{8}$ & 79.35$_{8}$ & 81.75$_{7}$ \\
    7 &  Think\_Positive  &  \checkmark  & 88.06$_{5}$ & 87.65$_{6}$ & 82.05$_{7}$ & 76.74$_{7}$ & 80.90$_{12}$ & 79.90$_{6}$ & 81.15$_{9}$ \\
    8 &  Kea$^{*}$  &   & 84.83$_{10}$ & 84.14$_{10}$ & 81.22$_{8}$ & 65.94$_{17}$ & 81.16$_{11}$ & 76.11$_{13}$ & 80.70$_{10}$ \\
    9 &  Lt\_{3}  &   & 86.28$_{8}$ & 85.26$_{7}$ & 81.02$_{9}$ & 70.76$_{13}$ & 80.44$_{13}$ & 77.41$_{11}$ & 80.33$_{13}$ \\
    10 &  senti.ue  &   & 84.05$_{11}$ & 78.72$_{16}$ & 80.54$_{10}$ & 82.75$_{1}$ & 81.90$_{6}$ & 81.73$_{5}$ & 81.47$_{8}$ \\
    11 &  LyS  &   & 85.69$_{9}$ & 81.44$_{12}$ & 79.92$_{11}$ & 71.67$_{10}$ & 83.95$_{4}$ & 78.51$_{10}$ & 82.21$_{5}$ \\
    12 &  UKPDIPF  &   & 80.45$_{15}$ & 79.05$_{14}$ & 79.67$_{12}$ & 65.63$_{18}$ & 81.42$_{9}$ & 75.57$_{14}$ & 80.33$_{11}$ \\
    13 &  UKPDIPF  &  \checkmark  & 80.45$_{16}$ & 79.05$_{15}$ & 79.67$_{13}$ & 65.63$_{19}$ & 81.42$_{10}$ & 75.57$_{15}$ & 80.33$_{12}$ \\
    14 &  TJP  &   & 81.13$_{14}$ & 84.41$_{9}$ & 79.30$_{14}$ & 71.20$_{12}$ & 78.27$_{15}$ & 76.26$_{12}$ & 78.39$_{15}$ \\
    15 &  SAP-RI  &   & 80.32$_{17}$ & 80.26$_{13}$ & 77.26$_{15}$ & 70.64$_{14}$ & 77.68$_{18}$ & 75.19$_{17}$ & 77.32$_{16}$ \\
    16 &  senti.ue$^{*}$  &  \checkmark  & 83.80$_{12}$ & 82.93$_{11}$ & 77.07$_{16}$ & 80.02$_{2}$ & 79.70$_{14}$ & 78.93$_{9}$ & 78.83$_{14}$ \\
    17 &  SAIL  &   & 78.47$_{18}$ & 74.46$_{20}$ & 76.89$_{17}$ & 65.56$_{20}$ & 70.62$_{22}$ & 71.02$_{21}$ & 72.57$_{21}$ \\
    18 &  columbia\_nlp$^\diamond$  &   & 81.50$_{13}$ & 74.55$_{19}$ & 76.54$_{18}$ & 61.76$_{22}$ & 78.19$_{16}$ & 72.16$_{19}$ & 77.11$_{18}$ \\
    19 &  IIT-Patna  &   & 76.54$_{20}$ & 75.99$_{18}$ & 76.43$_{19}$ & 71.43$_{11}$ & 77.99$_{17}$ & 75.28$_{16}$ & 77.26$_{17}$ \\
    20 &  Citius  &  \checkmark  & 76.59$_{19}$ & 69.31$_{21}$ & 75.21$_{20}$ & 68.40$_{15}$ & 75.82$_{20}$ & 73.14$_{18}$ & 75.38$_{19}$ \\
    21 &  Citius  &   & 74.71$_{21}$ & 61.44$_{25}$ & 73.03$_{21}$ & 65.18$_{21}$ & 71.64$_{21}$ & 69.95$_{22}$ & 71.90$_{22}$ \\
    22 &  IITPatna  &   & 70.91$_{23}$ & 77.04$_{17}$ & 72.25$_{22}$ & 66.32$_{16}$ & 76.03$_{19}$ & 71.53$_{20}$ & 74.45$_{20}$ \\
    23 &  SU-sentilab  &   & 74.34$_{22}$ & 62.58$_{24}$ & 68.26$_{23}$ & 53.31$_{25}$ & 69.53$_{23}$ & 63.70$_{24}$ & 68.59$_{23}$ \\
    24 &  Univ. Warwick$^{*}$  &   & 62.25$_{26}$ & 60.12$_{26}$ & 67.28$_{24}$ & 58.08$_{24}$ & 64.89$_{25}$ & 63.42$_{25}$ & 65.48$_{25}$ \\
    25 &  Univ. Warwick$^{*}$  &  \checkmark  & 64.91$_{25}$ & 63.01$_{23}$ & 67.17$_{25}$ & 60.59$_{23}$ & 67.46$_{24}$ & 65.07$_{23}$ & 67.14$_{24}$ \\
    26 &  DAEDALUS  &   & 67.42$_{24}$ & 63.92$_{22}$ & 60.98$_{26}$ & 45.27$_{27}$ & 61.01$_{26}$ & 55.75$_{26}$ & 60.50$_{26}$ \\
    27 &  DAEDALUS  &  \checkmark  & 61.95$_{27}$ & 55.97$_{27}$ & 58.11$_{27}$ & 49.19$_{26}$ & 58.65$_{27}$ & 55.32$_{27}$ & 58.17$_{27}$ \\
    \hline
     & Majority baseline  &  & 38.1 & 31.5 & 42.2 & 39.8 & 33.4 &  &\\
    \hline
    \end{tabular}
    \end{small}
    \caption{\label{tab:SubtaskA} \textbf{Results for subtask A.}
             The $^*$ indicates system resubmissions (because they initially trained on Twitter2013-test),
             and the $^\diamond$ indicates a system that includes a task co-organizer as a team member.
             The systems are sorted by their score on the Twitter2014 test dataset;
             the rankings on the individual datasets are indicated with a subscript.
             The last two columns show macro- and micro-averaged results across the three 2014 test datasets.}
  \end{table*}

Table~\ref{tab:SubtaskA} shows the results for subtask A, which attracted 27 submissions from 21 teams. There were seven unconstrained submissions: five teams submitted both a constrained and an unconstrained run, and two teams submitted an unconstrained run only. The best systems were constrained.
All participating systems outperformed the majority class baseline by a sizable margin.


\subsection{Subtask B}

\begin{table*}[t!]
    \centering
    \begin{small}
    \begin{tabular}{|c|l|c|cc|ccc|cc|}
    \hline
   & & \bf Uncon-& \multicolumn{2}{c|}{\bf 2013: Progress} & \multicolumn{3}{c|}{\bf 2014: Official} & \multicolumn{2}{c|}{\bf 2014: Average}\\
   \bf \# & \multicolumn{1}{|c|}{\bf System} & \bf strain.? & \bf Tweet & \bf SMS & \bf Tweet & \bf Tweet & \bf Live- & \bf Macro & \bf Micro\\
   & &  & &  & \bf & \bf sarcasm & \bf Journal & & \\
    \hline
    1 & TeamX  &  & 72.12$_{1}$ & 57.36$_{26}$ & 70.96$_{1}$ & 56.50$_{3}$ & 69.44$_{15}$ & 65.63$_{3}$ & 69.99$_{5}$ \\
    2 & coooolll  &  & 70.40$_{3}$ & 67.68$_{2}$ & 70.14$_{2}$ & 46.66$_{24}$ & 72.90$_{5}$ & 63.23$_{12}$ & 70.51$_{2}$ \\
    3 & RTRGO  &  & 69.10$_{5}$ & 67.51$_{3}$ & 69.95$_{3}$ & 47.09$_{23}$ & 72.20$_{6}$ & 63.08$_{13}$ & 70.15$_{3}$ \\
    4 & NRC-Canada  &  & 70.75$_{2}$ & 70.28$_{1}$ & 69.85$_{4}$ & 58.16$_{1}$ & 74.84$_{1}$ & 67.62$_{1}$ & 71.37$_{1}$ \\
    5 & TUGAS  &  & 65.64$_{13}$ & 62.77$_{11}$ & 69.00$_{5}$ & 52.87$_{12}$ & 69.79$_{13}$ & 63.89$_{6}$ & 68.84$_{8}$ \\
    6 & CISUC\_KIS$^{*}$  &  & 67.56$_{8}$ & 65.90$_{6}$ & 67.95$_{6}$ & 55.49$_{5}$ & 74.46$_{2}$ & 65.97$_{2}$ & 70.02$_{4}$ \\
    7 & SAIL  &  & 66.80$_{11}$ & 56.98$_{28}$ & 67.77$_{7}$ & 57.26$_{2}$ & 69.34$_{17}$ & 64.79$_{4}$ & 68.06$_{10}$ \\
    8 & SWISS-CHOCOLATE  &  & 64.81$_{18}$ & 66.43$_{5}$ & 67.54$_{8}$ & 49.46$_{16}$ & 73.25$_{4}$ & 63.42$_{10}$ & 69.15$_{6}$ \\
    9 & Synalp-Empathic  &  & 63.65$_{23}$ & 62.54$_{12}$ & 67.43$_{9}$ & 51.06$_{15}$ & 71.75$_{9}$ & 63.41$_{11}$ & 68.57$_{9}$ \\
    10 & Think\_Positive  & \checkmark  & 68.15$_{7}$ & 63.20$_{9}$ & 67.04$_{10}$ & 47.85$_{21}$ & 66.96$_{24}$ & 60.62$_{18}$ & 66.47$_{15}$ \\
    11 & SentiKLUE  &  & 69.06$_{6}$ & 67.40$_{4}$ & 67.02$_{11}$ & 43.36$_{30}$ & 73.99$_{3}$ & 61.46$_{14}$ & 68.94$_{7}$ \\
    12 & JOINT\_FORCES  & \checkmark  & 66.61$_{12}$ & 62.20$_{13}$ & 66.79$_{12}$ & 45.40$_{26}$ & 70.02$_{12}$ & 60.74$_{17}$ & 67.39$_{12}$ \\
    13 & AMI\_ERIC  &  & 70.09$_{4}$ & 60.29$_{20}$ & 66.55$_{13}$ & 48.19$_{20}$ & 65.32$_{26}$ & 60.02$_{21}$ & 65.58$_{20}$ \\
    14 & AUEB  &  & 63.92$_{21}$ & 64.32$_{8}$ & 66.38$_{14}$ & 56.16$_{4}$ & 70.75$_{11}$ & 64.43$_{5}$ & 67.71$_{11}$ \\
    15 & CMU-Qatar$^{*}$  &  & 65.11$_{17}$ & 62.95$_{10}$ & 65.53$_{15}$ & 40.52$_{38}$ & 65.63$_{25}$ & 57.23$_{27}$ & 64.87$_{24}$ \\
    16 & Lt\_3  &  & 65.56$_{14}$ & 64.78$_{7}$ & 65.47$_{16}$ & 47.76$_{22}$ & 68.56$_{20}$ & 60.60$_{19}$ & 66.12$_{17}$ \\
    17 & columbia\_nlp$^\diamond$  &  & 64.60$_{19}$ & 59.84$_{21}$ & 65.42$_{17}$ & 40.02$_{40}$ & 68.79$_{19}$ & 58.08$_{25}$ & 65.96$_{19}$ \\
    18 & LyS  &  & 66.92$_{10}$ & 60.45$_{19}$ & 64.92$_{18}$ & 42.40$_{33}$ & 69.79$_{14}$ & 59.04$_{22}$ & 66.10$_{18}$ \\
    19 & NILC\_USP  &  & 65.39$_{15}$ & 61.35$_{16}$ & 63.94$_{19}$ & 42.06$_{34}$ & 69.02$_{18}$ & 58.34$_{24}$ & 65.21$_{21}$ \\
    20 & senti.ue  &  & 67.34$_{9}$ & 59.34$_{23}$ & 63.81$_{20}$ & 55.31$_{6}$ & 71.39$_{10}$ & 63.50$_{7}$ & 66.38$_{16}$ \\
    21 & UKPDIPF  &  & 60.65$_{29}$ & 60.56$_{17}$ & 63.77$_{21}$ & 54.59$_{7}$ & 71.92$_{7}$ & 63.43$_{8}$ & 66.53$_{13}$ \\
    22 & UKPDIPF  & \checkmark  & 60.65$_{30}$ & 60.56$_{18}$ & 63.77$_{22}$ & 54.59$_{8}$ & 71.92$_{8}$ & 63.43$_{9}$ & 66.53$_{14}$ \\
    23 & SU-FMI$^{*}$$^\diamond$  &  & 60.96$_{28}$ & 61.67$_{15}$ & 63.62$_{23}$ & 48.34$_{19}$ & 68.24$_{21}$ & 60.07$_{20}$ & 64.91$_{23}$ \\
    24 & ECNU  &  & 62.31$_{27}$ & 59.75$_{22}$ & 63.17$_{24}$ & 51.43$_{14}$ & 69.44$_{16}$ & 61.35$_{15}$ & 65.17$_{22}$ \\
    25 & ECNU  & \checkmark  & 63.72$_{22}$ & 56.73$_{29}$ & 63.04$_{25}$ & 49.33$_{17}$ & 64.08$_{31}$ & 58.82$_{23}$ & 63.04$_{27}$ \\
    26 & Rapanakis  &  & 58.52$_{32}$ & 54.02$_{35}$ & 63.01$_{26}$ & 44.69$_{27}$ & 59.71$_{37}$ & 55.80$_{31}$ & 61.28$_{32}$ \\
    27 & Citius  & \checkmark  & 63.25$_{24}$ & 58.28$_{24}$ & 62.94$_{27}$ & 46.13$_{25}$ & 64.54$_{29}$ & 57.87$_{26}$ & 63.06$_{26}$ \\
    28 & CMUQ-Hybrid$^{*}$  &  & 63.22$_{25}$ & 61.75$_{14}$ & 62.71$_{28}$ & 40.95$_{37}$ & 65.14$_{27}$ & 56.27$_{30}$ & 63.00$_{28}$ \\
    29 & Citius  &  & 62.53$_{26}$ & 57.69$_{25}$ & 61.92$_{29}$ & 41.00$_{36}$ & 62.40$_{33}$ & 55.11$_{33}$ & 61.51$_{31}$ \\
    30 & KUNLPLab  &  & 58.12$_{33}$ & 55.89$_{31}$ & 61.72$_{30}$ & 44.60$_{28}$ & 63.77$_{32}$ & 56.70$_{29}$ & 62.00$_{29}$ \\
    31 & senti.ue$^{*}$  & \checkmark  & 65.21$_{16}$ & 56.16$_{30}$ & 61.47$_{31}$ & 54.09$_{9}$ & 68.08$_{22}$ & 61.21$_{16}$ & 63.71$_{25}$ \\
    32 & UPV-ELiRF  &  & 63.97$_{20}$ & 55.36$_{33}$ & 59.33$_{32}$ & 37.46$_{42}$ & 64.11$_{30}$ & 53.63$_{37}$ & 60.49$_{33}$ \\
    33 & USP\_Biocom  &  & 58.05$_{34}$ & 53.57$_{36}$ & 59.21$_{33}$ & 43.56$_{29}$ & 67.80$_{23}$ & 56.86$_{28}$ & 61.96$_{30}$ \\
    34 & DAEDALUS  & \checkmark  & 58.94$_{31}$ & 54.96$_{34}$ & 57.64$_{34}$ & 35.26$_{44}$ & 60.99$_{35}$ & 51.30$_{39}$ & 58.26$_{35}$ \\
    35 & IIT-Patna  &  & 52.58$_{40}$ & 51.96$_{37}$ & 57.25$_{35}$ & 41.33$_{35}$ & 60.39$_{36}$ & 52.99$_{38}$ & 57.97$_{36}$ \\
    36 & DejaVu  &  & 57.43$_{36}$ & 55.57$_{32}$ & 57.02$_{36}$ & 42.46$_{32}$ & 64.69$_{28}$ & 54.72$_{34}$ & 59.46$_{34}$ \\
    37 & GPLSI  &  & 57.49$_{35}$ & 46.63$_{42}$ & 56.06$_{37}$ & 53.90$_{10}$ & 57.32$_{41}$ & 55.76$_{32}$ & 56.47$_{37}$ \\
    38 & BUAP  &  & 56.85$_{37}$ & 44.27$_{44}$ & 55.76$_{38}$ & 51.52$_{13}$ & 53.94$_{44}$ & 53.74$_{36}$ & 54.97$_{39}$ \\
    39 & SAP-RI  &  & 50.18$_{44}$ & 49.00$_{41}$ & 55.47$_{39}$ & 48.64$_{18}$ & 57.86$_{40}$ & 53.99$_{35}$ & 56.17$_{38}$ \\
    40 & UMCC\_DLSI\_Sem  &  & 51.96$_{41}$ & 50.01$_{38}$ & 55.40$_{40}$ & 42.76$_{31}$ & 53.12$_{45}$ & 50.43$_{40}$ & 54.20$_{42}$ \\
    41 & IBM\_EG  &  & 54.51$_{38}$ & 46.62$_{43}$ & 52.26$_{41}$ & 34.14$_{46}$ & 59.24$_{38}$ & 48.55$_{43}$ & 54.34$_{41}$ \\
    42 & Alberta  &  & 53.85$_{39}$ & 49.05$_{40}$ & 52.06$_{42}$ & 40.40$_{39}$ & 52.38$_{46}$ & 48.28$_{44}$ & 51.85$_{44}$ \\
    43 & lsis\_lif  &  & 46.38$_{46}$ & 38.56$_{47}$ & 52.02$_{43}$ & 34.64$_{45}$ & 61.09$_{34}$ & 49.25$_{41}$ & 54.90$_{40}$ \\
    44 & SU-sentilab  &  & 50.17$_{45}$ & 49.60$_{39}$ & 49.52$_{44}$ & 31.49$_{47}$ & 55.11$_{42}$ & 45.37$_{47}$ & 51.09$_{45}$ \\
    45 & SINAI  &  & 50.59$_{42}$ & 57.34$_{27}$ & 49.50$_{45}$ & 31.15$_{49}$ & 58.33$_{39}$ & 46.33$_{46}$ & 52.26$_{43}$ \\
    46 & IITPatna  &  & 50.32$_{43}$ & 40.56$_{46}$ & 48.22$_{46}$ & 36.73$_{43}$ & 54.68$_{43}$ & 46.54$_{45}$ & 50.29$_{46}$ \\
    47 & Univ. Warwick  &  & 39.17$_{48}$ & 29.50$_{49}$ & 45.56$_{47}$ & 39.77$_{41}$ & 39.60$_{49}$ & 41.64$_{48}$ & 43.19$_{48}$ \\
    48 & UMCC\_DLSI\_Graph  &  & 43.24$_{47}$ & 36.66$_{48}$ & 45.49$_{48}$ & 53.15$_{11}$ & 47.81$_{47}$ & 48.82$_{42}$ & 46.56$_{47}$ \\
    49 & Univ. Warwick  & \checkmark  & 34.23$_{50}$ & 24.63$_{50}$ & 45.11$_{49}$ & 31.40$_{48}$ & 29.34$_{50}$ & 35.28$_{49}$ & 38.88$_{49}$ \\
    50 & DAEDALUS  &  & 36.57$_{49}$ & 40.86$_{45}$ & 33.03$_{50}$ & 28.96$_{50}$ & 40.83$_{48}$ & 34.27$_{50}$ & 35.81$_{50}$ \\
    \hline
     & Majority baseline  &  & 29.2 & 19.0 & 34.6 & 27.7 & 27.2 &  &\\
    \hline
    \end{tabular}
    \end{small}
    \caption{\label{tab:SubtaskB} \textbf{Results for subtask B.}
             The $^*$ indicates system resubmissions (because they initially trained on Twitter2013-test),
             and the $^\diamond$ indicates a system that includes a task co-organizer as a team member.
             The systems are sorted by their score on the Twitter2014 test dataset;
             the rankings on the individual datasets are indicated with a subscript.
             The last two columns show macro- and micro-averaged results across the three 2014 test datasets.}
  \end{table*}

The results for subtask B are shown in Table~\ref{tab:SubtaskB}. The subtask attracted 50 submissions from 44 teams. There were eight unconstrained submissions: six teams submitted both a constrained and an unconstrained run, and two teams submitted an unconstrained run only. As for subtask A, the best systems were constrained. Again, all participating systems outperformed the majority class baseline; however, some systems were very close to it.

\section{Discussion}

Overall, we observed similar trends as in SemEval-2013 Task 2.
Almost all systems used supervised learning.
Most systems were constrained, including the best ones in all categories.
As in 2013, we observed several cases of a team submitting a constrained
and an unconstrained run and the constrained run performing better.

It is unclear why unconstrained systems did not outperform constrained ones.
It could be because participants did not use enough external data
or because the data they used was too different from Twitter or from our annotation method.
Or it could be due to our definition of \emph{unconstrained},
which labels as unconstrained systems that use additional tweets directly,
but considers unconstrained those that use additional tweets to build sentiment lexicons and then use these lexicons.

As in 2013, the most popular classifiers were SVM, MaxEnt, and Naive Bayes.
Moreover, two submissions used deep learning,
\emph{coooolll} (Harbin Institute of Technology) and \emph{ThinkPositive} (IBM Research, Brazil),
which were ranked second and tenth on subtask B, respectively.

The features used were quite varied, including word-based
(e.g.,~word and character $n$-grams, word shapes, and lemmata),
syntactic, and Twitter-specific such as emoticons and abbreviations.
The participants still relied heavily on lexicons of opinion words,
the most popular ones being the same as in 2013: MPQA, SentiWordNet and Bing Liu's opinion lexicon.
Popular this year was also the NRC lexicon \cite{MohammadKZ2013},
created by the best-performing team in 2013,
which is top-performing this year as well.

Preprocessing of tweets was still a popular technique. In addition to standard NLP steps such as tokenization, stemming, lemmatization, stop-word removal and POS tagging, most teams applied some kind of Twitter-specific processing such as substitution/removal of URLs, substitution of emoticons, word normalization, abbreviation lookup, and punctuation removal. Finally, several of the teams used Twitter-tuned NLP tools such as part of speech and named entity taggers \cite{gimpel2011,ner}.

The similarity of preprocessing techniques, NLP tools, classifiers and features used
in 2013 and this year is probably partially due to many teams participating in both years.
As Table~\ref{T:teams} shows, 18 out of the 46 teams are returning teams.

Comparing the results on the progress Twitter test in 2013 and 2014, we can see that \emph{NRC-Canada}, the 2013 winner for subtask A,
have now improved their F1 score from 88.93 to 90.14, which is the 2014 best score.
The best score on the Progress SMS in 2014 of 89.31 belongs to \emph{ECNU}; this is a big jump compared to their 2013 score of 76.69,
but it is less compared to the 2013 best of 88.37 achieved by \emph{GU-MLT-LT}.
For subtask B, on the Twitter progress testset,
the 2013 winner \emph{NRC-Canada} improves their 2013 result from 69.02 to 70.75, which is the second best in 2014;
the winner in 2014, \emph{TeamX}, achieves 72.12.
On the SMS progress test, the 2013 winner \emph{NRC-Canada} improves its F1 score from 68.46 to 70.28.
Overall, we see consistent improvements on the progress testset for both subtasks: 0-1 and 2-3 points absolute for subtasks A and B, respectively.

Finally, note that for both subtasks, the best systems on the Twitter-2014 dataset are those that performed best on the 2013 progress Twitter dataset: \emph{NRC-Canada} for subtask A, and \emph{TeamX} (Fuji Xerox Co., Ltd.) for subtask B.

It is interesting to note that the best results for Twitter2014-test are lower than those for Twitter2013-test
for both subtask A (86.63 vs. 90.14) and subtask B (70.96 vs 72.12).
This is so despite the baselines for Twitter2014-test being higher than those for Twitter2013-test:
42.2 vs. 38.1 for subtask A, and 34.6 vs. 29.2 for subtask B.
Most likely, having access to Twitter2013-test at development time, teams have overfitted on it.
It could be also the case that some of the sentiment dictionaries that were built in 2013
have become somewhat outdated by 2014.

Finally, note that while some teams such as \emph{NRC-Canada} performed well across all test sets,
other such as \emph{TeamX}, which used a weighting scheme tuned specifically for class imbalances in tweets,
were only strong on Twitter datasets.

\section{Conclusion}

We have described the data, the experimental setup and the results for SemEval-2014 Task 9. As in 2013, our task was the most popular one at SemEval-2014, attracting 46 participating teams: 21 in subtask A (27 submissions) and 44 in subtask B (50 submissions).


We introduced three new test sets for 2014:
an in-domain Twitter dataset,
an out-of-domain LiveJournal test set,
and a dataset of tweets containing sarcastic content.
While the performance on the LiveJournal test set was mostly comparable to the in-domain Twitter test set,
for most teams there was a sharp drop in performance for sarcastic tweets,
highlighting better handling of sarcastic language as one important direction for future work in Twitter sentiment analysis.

We plan to run the task again in 2015 with the inclusion of a new sub-evaluation on detecting sarcasm
with the goal of stimulating research in this area; we further plan to add one more test domain.

In the 2015 edition of the task, we might also remove the constrained/unconstrained distinction.

Finally, as there are multiple opinions about a topic in Twitter,
we would like to focus on detecting the sentiment trend towards a topic.


\section*{Acknowledgements}

We would like to thank Kathleen McKeown and Smaranda Muresan for funding the 2014 Twitter test sets.
We also thank the anonymous reviewers.

\afterpage{
\begin{table*}[ht!]
\small
\begin{center}
\begin{tabular}{|c|l|l|c|}
\hline
  \bf Subtasks & \multicolumn{1}{c|}{\bf Team} & \multicolumn{1}{c|}{\bf Affiliation} & \bf 2013?\\
  \hline
  B & Alberta & University of Alberta &\\
  B & AMI\_ERIC & AMI Software R\&D and Universit\'{e} de Lyon (ERIC LYON 2) & yes\\
  B & AUEB & Athens University of Economics and Business & yes\\
  B & BUAP & Benem\'{e}rita Universidad Aut\'{o}noma de Puebla &\\
  B & CISUC\_KIS & University of Coimbra &\\
  A, B & Citius & University of Santiago de Compostela &\\
  A, B & CMU-Qatar & Carnegie Mellon University, Qatar &\\
  A, B & CMUQ-Hybrid & Carnegie Mellon University, Qatar (different from the above) &\\
  A, B & columbia\_nlp & Columbia University & yes\\
  B & cooolll & Harbin Institute of Technology &\\
  A, B & DAEDALUS & Daedalus &\\
  B & DejaVu & Indian Institute of Technology, Kanpur &\\
  A, B & ECNU & East China Normal University & yes\\
  B & GPLSI & University of Alicante &\\
  B & IBM\_EG & IBM Egypt &\\
  A, B & IITPatna & Indian Institute of Technology, Patna &\\
  A, B & IIT-Patna & Indian Institute of Technology, Patna (different from the above) &\\
  B & JOINT\_FORCES & Zurich University of Applied Sciences &\\
  A & Kea & York University, Toronto & yes\\
  B & KUNLPLab & Ko\c{c} University &\\
  B & lsis\_lif & Aix-Marseille University & yes\\
  A, B & Lt\_3 & Ghent University &\\
  A, B & LyS & Universidade da Coru\~{n}a &\\
  B & NILC\_USP & University of S\~{a}o Paulo & yes\\
  A, B & NRC-Canada & National Research Council Canada & yes\\
  B & Rapanakis & Stamatis Rapanakis &\\
  B & RTRGO & Retresco GmbH and University of Gothenburg & yes\\
  A, B & SAIL & Signal Analysis and Interpretation Laboratory & yes\\
  A, B & SAP-RI & SAP Research and Innovation &\\
  A, B & senti.ue & Universidade de \'{E}vora & yes\\
  A, B & SentiKLUE & Friedrich-Alexander-Universit\"{a}t Erlangen-N\"{u}rnberg & yes\\
  B & SINAI & University of Ja\'{e}n & yes\\
  B & SU-FMI & Sofia University &\\
  A, B & SU-sentilab & Sabanci University & yes\\
  B & SWISS-CHOCOLATE & ETH Zurich &\\
  B & Synalp-Empathic & University of Lorraine &\\
  B & TeamX & Fuji Xerox Co., Ltd. &\\
  A, B & Think\_Positive & IBM Research, Brazil &\\
  A & TJP & University of Northumbria at Newcastle Upon Tyne & yes\\
  B & TUGAS & Instituto de Engenharia de Sistemas e Computadores, & yes\\
  & & Investiga\c{c}\~{a}o e Desenvolvimento em Lisboa &  \\
  A, B & UKPDIPF & Ubiquitous Knowledge Processing Lab &\\
  B & UMCC\_DLSI\_Graph & Universidad de Matanzas and Univarsidad de Alicante & yes\\
  B & UMCC\_DLSI\_Sem & Universidad de Matanzas and Univarsidad de Alicante  (different from above) & yes\\
  A, B & Univ. Warwick & University of Warwick &\\
  B & UPV-ELiRF & Universitat Polit\`{e}cnica de Val\`{e}ncia &\\
  B & USP\_Biocom & University of S\~{a}o Paulo and Federal University of S\~{a}o Carlos &\\
\hline
\end{tabular}
\caption{Participating teams, their affiliations, subtasks they have taken part in,
         and an indication about whether the team participated in SemEval-2013 Task 2.}
\label{T:teams}
\end{center}
\end{table*}
\clearpage}

\bibliographystyle{acl}
\bibliography{acl2014}

\begin{thebibliography}{}

\bibitem[\protect\citename{Baccianella \bgroup et al.\egroup }2010]{swn}
Stefano Baccianella, Andrea Esuli, and Fabrizio Sebastiani.
\newblock 2010.
\newblock {SentiWordNet} 3.0: An enhanced lexical resource for sentiment
  analysis and opinion mining.
\newblock In {\em Proceedings of the Seventh International Conference on
  Language Resources and Evaluation}, LREC~'10, Valletta, Malta.

\bibitem[\protect\citename{Barbosa and Feng}2010]{Barbosa10}
Luciano Barbosa and Junlan Feng.
\newblock 2010.
\newblock Robust sentiment detection on {T}witter from biased and noisy data.
\newblock In {\em Proceedings of the 23rd International Conference on
  Computational Linguistics: Posters}, COLING '10, pages 36--44, Beijing,
  China.

\bibitem[\protect\citename{Bifet \bgroup et al.\egroup }2011]{Bifet11}
Albert Bifet, Geoffrey Holmes, Bernhard Pfahringer, and Ricard Gavald{\`a}.
\newblock 2011.
\newblock Detecting sentiment change in {T}witter streaming data.
\newblock {\em Journal of Machine Learning Research, Proceedings Track},
  17:5--11.

\bibitem[\protect\citename{Chen and Kan}2013]{SMScorpus}
Tao Chen and Min-Yen Kan.
\newblock 2013.
\newblock Creating a live, public short message service corpus: the {NUS SMS}
  corpus.
\newblock {\em Language Resources and Evaluation}, 47(2):299--335.

\bibitem[\protect\citename{Davidov \bgroup et al.\egroup }2010]{Davidov10}
Dmitry Davidov, Oren Tsur, and Ari Rappoport.
\newblock 2010.
\newblock Semi-supervised recognition of sarcasm in {T}witter and {A}mazon.
\newblock In {\em Proceedings of the Fourteenth Conference on Computational
  Natural Language Learning}, CoNLL '10, pages 107--116, Uppsala, Sweden.

\bibitem[\protect\citename{Gimpel \bgroup et al.\egroup }2011]{gimpel2011}
Kevin Gimpel, Nathan Schneider, Brendan O'Connor, Dipanjan Das, Daniel Mills,
  Jacob Eisenstein, Michael Heilman, Dani Yogatama, Jeffrey Flanigan, and
  Noah~A. Smith.
\newblock 2011.
\newblock Part-of-speech tagging for {T}witter: Annotation, features, and
  experiments.
\newblock In {\em Proceedings of the 49th Annual Meeting of the Association for
  Computational Linguistics: Human Language Technologies}, ACL-HLT '11, pages
  42--47, Portland, Oregon, USA.

\bibitem[\protect\citename{Gonz\'alez-Ib\'a\~nez \bgroup et al.\egroup
  }2011]{conf/acl/Gonzalez-IbanezMW11}
Roberto Gonz\'alez-Ib\'a\~nez, Smaranda Muresan, and Nina Wacholder.
\newblock 2011.
\newblock Identifying sarcasm in {T}witter: a closer look.
\newblock In {\em Proceedings of the 49th Annual Meeting of the Association for
  Computational Linguistics: Human Language Technologies - Short Papers},
  ACL-HLT '11, pages 581--586, Portland, Oregon, USA.

\bibitem[\protect\citename{Jansen \bgroup et al.\egroup }2009]{Jansen09}
Bernard Jansen, Mimi Zhang, Kate Sobel, and Abdur Chowdury.
\newblock 2009.
\newblock Twitter power: Tweets as electronic word of mouth.
\newblock {\em J. Am. Soc. Inf. Sci. Technol.}, 60(11):2169--2188.

\bibitem[\protect\citename{Kouloumpis \bgroup et al.\egroup
  }2011]{Kouloumpis11}
Efthymios Kouloumpis, Theresa Wilson, and Johanna Moore.
\newblock 2011.
\newblock Twitter sentiment analysis: The good the bad and the {OMG}!
\newblock In {\em Proceedings of the Fifth International Conference on Weblogs
  and Social Media}, ICWSM '11, Barcelona, Catalonia, Spain.

\bibitem[\protect\citename{Liebrecht \bgroup et al.\egroup
  }2013]{liebrecht-kunneman-vandenbosch:2013:WASSA}
Christine Liebrecht, Florian Kunneman, and Antal Van~den Bosch.
\newblock 2013.
\newblock The perfect solution for detecting sarcasm in tweets \#not.
\newblock In {\em Proceedings of the 4th Workshop on Computational Approaches
  to Subjectivity, Sentiment and Social Media Analysis}, pages 29--37, Atlanta,
  Georgia, USA.

\bibitem[\protect\citename{Mohammad \bgroup et al.\egroup
  }2013]{MohammadKZ2013}
Saif Mohammad, Svetlana Kiritchenko, and Xiaodan Zhu.
\newblock 2013.
\newblock {NRC-Canada}: Building the state-of-the-art in sentiment analysis of
  tweets.
\newblock In {\em Proceedings of the Seventh international workshop on Semantic
  Evaluation Exercises}, SemEval-2013, pages 321--327, Atlanta, Georgia, USA.

\bibitem[\protect\citename{Nakov \bgroup et al.\egroup }2013]{semeval2013}
Preslav Nakov, Sara Rosenthal, Zornitsa Kozareva, Veselin Stoyanov, Alan
  Ritter, and Theresa Wilson.
\newblock 2013.
\newblock {SemEval}-2013 task 2: Sentiment analysis in {T}witter.
\newblock In {\em Second Joint Conference on Lexical and Computational
  Semantics (*SEM), Volume 2: Proceedings of the Seventh International Workshop
  on Semantic Evaluation}, SemEval '13, pages 312--320, Atlanta, Georgia, USA.

\bibitem[\protect\citename{O'Connor \bgroup et al.\egroup }2010]{oconnor10}
Brendan O'Connor, Ramnath Balasubramanyan, Bryan Routledge, and Noah Smith.
\newblock 2010.
\newblock From tweets to polls: Linking text sentiment to public opinion time
  series.
\newblock In {\em Proceedings of the Fourth International Conference on Weblogs
  and Social Media}, ICWSM~'10, Washington, DC, USA.

\bibitem[\protect\citename{Pak and Paroubek}2010]{Pak10}
Alexander Pak and Patrick Paroubek.
\newblock 2010.
\newblock Twitter based system: Using {T}witter for disambiguating sentiment
  ambiguous adjectives.
\newblock In {\em Proceedings of the 5th International Workshop on Semantic
  Evaluation}, SemEval '10, pages 436--439, Uppsala, Sweden.

\bibitem[\protect\citename{Pang \bgroup et al.\egroup }2002]{Pang:2002:TUS}
Bo~Pang, Lillian Lee, and Shivakumar Vaithyanathan.
\newblock 2002.
\newblock Thumbs up?: Sentiment classification using machine learning
  techniques.
\newblock In {\em Proceedings of the Conference on Empirical Methods in Natural
  Language Processing - Volume 10}, EMNLP '02, pages 79--86.

\bibitem[\protect\citename{Pontiki \bgroup et al.\egroup
  }2014]{Semeval2014task4}
Maria Pontiki, Harris Papageorgiou, Dimitrios Galanis, Ion Androutsopoulos,
  John Pavlopoulos, and Suresh Manandhar.
\newblock 2014.
\newblock {SemEval}-2014 task 4: Aspect based sentiment analysis.
\newblock In {\em Proceedings of the 8th International Workshop on Semantic
  Evaluation}, SemEval '14, Dublin, Ireland.

\bibitem[\protect\citename{Ritter \bgroup et al.\egroup }2011]{ner}
Alan Ritter, Sam Clark, Mausam, and Oren Etzioni.
\newblock 2011.
\newblock Named entity recognition in tweets: An experimental study.
\newblock In {\em Proceedings of the Conference on Empirical Methods in Natural
  Language Processing}, EMNLP '11, pages 1524--1534, Edinburgh, Scotland, UK.

\bibitem[\protect\citename{Tumasjan \bgroup et al.\egroup }2010]{Tumasjan10}
Andranik Tumasjan, Timm Sprenger, Philipp Sandner, and Isabell Welpe.
\newblock 2010.
\newblock Predicting elections with {T}witter: What 140 characters reveal about
  political sentiment.
\newblock In {\em Proceedings of the Fourth International Conference on Weblogs
  and Social Media}, ICWSM~'10, Washington, DC, USA.

\bibitem[\protect\citename{Wiebe \bgroup et al.\egroup }2005]{Wiebe05}
Janyce Wiebe, Theresa Wilson, and Claire Cardie.
\newblock 2005.
\newblock Annotating expressions of opinions and emotions in language.
\newblock {\em Language Resources and Evaluation}, 39(2-3):165--210.

\end{thebibliography}

\end{document}